\documentclass[journal,comsoc]{IEEEtran}
\UseRawInputEncoding
\usepackage{amsmath}
\usepackage{bm} 
\usepackage{mathrsfs}
\usepackage{amssymb}
\usepackage{subfigure}
\usepackage{algorithm}
\usepackage{algorithmic}
\usepackage{graphicx}
\usepackage{comment}
\usepackage{color}
\usepackage{stfloats}
\usepackage{multirow}
\usepackage{graphicx}
\usepackage{caption}
\usepackage{graphicx, subfig}
\usepackage{threeparttable}
\usepackage{verbatim}
\usepackage{booktabs}
\usepackage{dutchcal}

\begin{document}
\title{Multi-Agent Broad Reinforcement Learning for Intelligent Traffic Light Control}

\author{{Ruijie Zhu,~\IEEEmembership{Member,~IEEE,}
		Lulu Li,~\IEEEmembership{Student Member,~IEEE,}
		Shuning Wu,~\IEEEmembership{Student Member,~IEEE,}\\
		Pei Lv,~\IEEEmembership{Member,~IEEE,}
		Yafai Li,~\IEEEmembership{Member,~IEEE,}
		and Mingliang Xu,~\IEEEmembership{Member,~IEEE}}
	\thanks{
		This preprint is intended for publication in a scientific journal, but has not been peer-reviewed. Copyright and all rights therein are maintained by the Authors or by other copyright owners. It is understood that all persons copying this information will adhere to the terms and constraints invoked by each Author?s copyright.
}}

\markboth{}
{Shell \MakeLowercase{\textit{et al.}}: Bare Demo of IEEEtran.cls for IEEE Journals}
\maketitle

\begin{abstract}
Intelligent Traffic Light Control System (ITLCS) is a typical Multi-Agent System (MAS), which comprises multiple roads and traffic lights. Constructing a model of MAS for ITLCS is the basis to alleviate traffic congestion.
Existing approaches of MAS are largely based on Multi-Agent Deep Reinforcement Learning (MADRL). 
Although the Deep Neural Network (DNN) of MABRL is effective, the training time is long, and the parameters are difficult to trace.  
Recently, Broad Learning Systems (BLS) provided a selective way for learning in the deep neural networks by a flat network. Moreover, Broad Reinforcement Learning (BRL) extends BLS in Single Agent Deep Reinforcement Learning (SADRL) problem with promising results. 
However, BRL does not focus on the intricate structures and interaction of agents. Motivated by the feature of MADRL and the issue of BRL, we propose a Multi-Agent Broad Reinforcement Learning (MABRL) framework to explore the function of BLS in MAS.
Firstly, unlike most existing MADRL approaches, which use a series of deep neural networks structures, we model each agent with broad networks. Then, we introduce a dynamic self-cycling interaction mechanism to confirm the "3W" information: When to interact, Which agents need to consider, What information to transmit.
Finally, we do the experiments based on the intelligent traffic light control scenario. We compare the MABRL approach with six different approaches, and experimental results on three datasets verify the effectiveness of MABRL.
\end{abstract}


\section{Introduction}
Intelligent Traffic Light Control (ITLC) is a typical application of Multi-Agent Systems (MAS) \cite{tits_tlc2}. Intersections of road networks in ITLC are modeled as agents, and agents aim to relieve traffic congestion \cite{tiv_tlc1}. Therefore, building the suitable MAS model of ITLC is the basis for achieving goals. 
In MAS, agents are a series of intelligence controllers with autonomy, inferential capability, and social behavior. MAS organize a group of agents to achieve a common objective by interacting, making decisions, coordinating, and learning \cite{MASintroduction}. 
While MAS has significant progress in many areas \cite{bc2,visualization}, MAS is still facing many challenges. 
Firstly, each agent of MAS is a local observer and has limited perception. 
Secondly, the decisions of each agent will disturb the whole environment, causing the environment becomes unstable. Thus, the mappings from state to action among Multiple Agents (MA) are complicated. 
Thirdly, the interaction of different agents needs to consider. Aiming to achieve global optimal resolution, agents require to quantify the information and ability about other agents, then build suitable inference mechanisms \cite{2018survery, survery2021} to make decisions. 
In summary, agents of MAS need to take account of perception, decision, and inference.

Many approaches have been proposed to enhance the performance of MAS \cite{ACO, MH,PSO,tiv_madrl_1}. Among these methods, MADRL receives more attention. MADRL adopts the Deep Reinforcement Learning (DRL) algorithm with Deep Neural Network (DNN) to solve problems of MAS. 
Comparing the typical structure of traditional Single-Agent DRL (SADRL), the MADRL approach adds a new section to ponder the influence of other agents. There are two reasons for the MADRL approaches not only the simple extension of SADRL.
In terms of information in MADRL, the information space scales up as the number of agents increases. Both the own and joint state-action values of individual agents should be utilized. Thus, storage pressure and the calculation difficulty are heavier than SADRL.
In terms of the framework in MADRL, Fully Decentralized (FD) \cite{FD}, Fully Centralized (FC) \cite{FC}, Centralized Training and Decentralized Execution (CTDE) \cite{CTDE} are typically paradigms of MADRL, they all constituted by DNN. 
Before agents make the decision, these paradigms of MADRL standardize the interaction structure by constructing serviceable and multiple layers.
As traditional SADRL approaches without the cross and parallelism of policy between MA, the network structures of MADRL are more intricate than SADRL, and the training time will be lengthened when the parameters transmit layers by layers.
Accordingly, MADRL has a heavier calculation burden and intricate structure than SADRL.

Broad Learning Systems (BLS) \cite{2017Broad}, \cite{BLS} is an incremental algorithm inspired by the Single-Layer Feedforward Neural networks (SLFN) \cite{SLFN}. BLS finds a new way with fast remodeling speed to substitute the learning of DNN. Unlike the single layer network structure, such as Radial Basis Function (RBF) \cite{RBF}, BLS uses the mapped nodes and enhancement nodes to handle the input data and trains the models regarding ridge regression.
Besides, Broad Learning with RL signal Feedback (BLRLF) \cite{BLRLF} exploits BLS to RL area by introducing a weight optimization mechanism into Adaptive Dynamic Programming (ADP) \cite{ADP} to enhance the expansion capability of BLS. 
Recently, Broad Reinforcement Learning (BRL) \cite{IOT-BRL} has been investigated combining the BLS and DRL.
The framework of BRL has two important keys, one is using the Broad Networks (BN) in BLS to replace the DNN, and the other is adopting the training pool to introduce the labels of BLS. Compared with DRL approaches, BRL has better performance with a shorter execution time. 
BRL is the first algorithm to solve the control questions by using the BLS.
However, BRL has concentrated on the problem with single agent without paying little attention to the MAS. With the number of agents increasing, it is worth investigating how BLS handles the mutual effect between MA.

Inspired by the issues of MADRL and the feature of BRL, we propose Multi-Agent Broad Reinforcement Learning (MABRL). 
Firstly, we outline the framework of MABRL, which combines the MADRL and BRL to explore the function of BLS in MAS. Each agent has integrated decision-making structures with broad networks, and updates policy based on the memory and pseudoinverse calculation.
Secondly, agents of MABRL adopt the joint policy based on the stochastic game to interact with the environment continually. They evaluate the influences of other agents by the Dynamic Self-Cycling Interaction Mechanism (DSCIM) to make decisions. 
Finally, we model an instance of ITLC and experiment with three datasets to verify the effectiveness of MABRL.
To the best of our knowledge, this is the first approach that applies BRL in MADRL. The contributions of our work summarize as follow:

\begin{itemize}
	  \item [(1)] A novel MABRL framework has been proposed adopting the BRL to solve the problems of MAS. Unlike the traditional MADRL, MABRL has a simple and traceable BN structure and updates training models and parameters by pseudoinverse calculation. Compared with BRL, multiple agents of MABRL adopt BN with interaction mechanisms to make decisions.
	  
	  \item [(2)]The Dynamic Self-Cycling Interaction Mechanism (DSCIM) has been designed in MABRL to enhance the interaction between agents, which accounts for the attention mechanism. Agents adopt DSCIM to confirm the joint information about "3W": When to interact with others, Which agents need to consider, and What informations need to transmit. After obtaining the joint information, agents of MABRL conduct the mapped features and joint information by enhancement nodes.
	  
	  \item [(3)] We build a model of ITLC with MABRL. Three datasets are considered to experiment. The ability of MABRL can be measured by relieving traffic congestion when the environment and datasets are intricate.
	  
\end{itemize}

\section{Conclusion} \label{con}
In this paper, we propose a Multi-Agent Broad Reinforcement Learning (MABRL) framework for developing the interaction of MA with broad networks.
Specifically, BLS is used to replace the network architecture of MADRL with DNN, making the approach more flexible and intelligible. Besides, we design DSCIM inspired by AM to enhance the interaction between agents. Moreover, we apply the MABRL approach in ITLC to adjust the traffic flow dynamically. The results of the experiments manifest the MABRL approach obtains stable performance of different scenarios compared with other approaches for ITLC.
In the future, we will focus on more complex structures and paradigms of MABRL to solve the problems of MAS in a new way, and apply the MABRL in more practical and extensive scenarios.

\bibliographystyle{IEEEtran}
\bibliography{mabrl_2022_arvix.bib}

\end{document}